\documentclass[]{spie}  
\usepackage[]{graphicx}
\usepackage{amsmath}
\usepackage{subfigure}
\usepackage{booktabs}
\usepackage[raggedright]{titlesec}
\usepackage{algorithm}
\usepackage{enumerate}
\usepackage{algorithmic}
\usepackage{subfigure}
\newtheorem{example}{Example}[section]
\title{An Adaptive Amoeba Algorithm for Shortest Path Tree Computation in Dynamic Graphs}

\author{Xiaoge Zhang \supit{a}\supit{,}\supit{f}, Qi Liu\supit{b}\supit{,}\supit{c}\supit{,}\supit{g}, Yong Hu\supit{d}, Felix T. S. Chan\supit{e}, Sankaran Mahadevan\supit{f}, Zili Zhang\supit{a}\supit{,}\supit{g}, Yong Deng \supit{a}\supit{,}\supit{f}\supit{*}
\skiplinehalf
\supit{a}School of Computer and Information Science, Southwest University, Chongqing 400715, China;
\supit{b}Department of Biomedical Informatics, Medical Center, Vanderbilt University, Nashiville, 37235, USA;
\supit{c}School of Life Sciences and Biotechnology, Shanghai Jiao Tong University, Shanghai, 200030, China;
\supit{d}Institute of Business Intelligence and Knowledge Discovery, Guangdong University of Foreign Studies, Guangzhou 510006, China;
\supit{e}Department of Industrial and Systems Engineering, The Hong Kong Polytechnic University, Hung Hum, Kowloon, Hong Kong
\supit{f}School of Engineering, Vanderbilt University, Nashiville, 37235, USA
\supit{g}These authors contribute equally\\
\supit{*}Correspondence and requests for materials should be addressed to Y. D. (prof.deng@hotmail.com)
}

\begin{document}
  \maketitle

This paper presents an adaptive amoeba algorithm to address the shortest path tree (SPT) problem in dynamic graphs. In dynamic graphs, the edge weight updates consists of three categories: edge weight increases, edge weight decreases, the mixture of them. Existing work on this problem solve this issue through analyzing the nodes influenced by the edge weight updates and recompute these affected vertices. However, when the network becomes big, the process will become complex. The proposed method can overcome the disadvantages of the existing approaches. The most important feature of this algorithm is its adaptivity. When the edge weight changes, the proposed algorithm can recognize the affected vertices and reconstruct them spontaneously. To evaluate the proposed adaptive amoeba algorithm, we compare it with the Label Setting algorithm and Bellman-Ford algorithm. The comparison results demonstrate the effectiveness of the proposed method.


\section*{}
The shortest path tree problem (SPT) is one of the basic network optimization problems and it is a variance of the shortest path problem (SPP), which has been widely used in many fields, such as multicast routing \cite{bauer1996distributed}, Route Information Protocol (RIP) \cite{rescigno2001optimally}, wireless network \cite{zhao2012bounded}, network design \cite{wang2012understanding,barthelemy2013self}, IS-IS \cite{perlman1991comparison}, and complex networks \cite{estrada2013peer,wei2013box,szell2012understanding}. Its objective is to find the set of edges which connect all the nodes in the network so that the sum of the edge lengths from the source to the other nodes is minimized. As the SPT problem is frequently as subproblem when solving many combinatorial and network optimization problems, many researchers payed their attention to this problem \cite{chan2009shortest,chen2010low}.

In practical environment, the network is changing with time. It lead to the occurence of the variance of SPTs named Dynamic Shortest Path (DSP) problem. Assume $G\left( {V,E,\omega } \right)$ be a simple network and the edge weights in this network are nonnegative numbers. Let ${G^{'}}\left( {V,E,{\omega ^{'}}} \right)$ be another network obtained from $G$ in which some edge weights change. Suppose $T_s$ and $T_s^{'}$ are the SPTs rooted at $s$ in $G$ and $G^{'}$. The DSP problem is to compute $T_s^{'}$ from $T_s$. Many methods have been proposed to deal with this problem. For example, the classical Dijkstra algorithm \cite{Dijksra1959} and Bellman algorithm solve this problem through recalculating the SPTs whenever there is a change to edge weights. Due to the high computational time, it cannot meet the requirement of the emergent accidents. After the idea of using an SPTs update program which only reconstruct on the affected vertices proposed by Frigioni \cite{frigioni1994incremental} appeared, many dynamic SPTs approaches have implemented this concept into real-world applications \cite{chan2009shortest,chen2010low} to reduce the computational time. They have divided the edge weight updates into three categories: edge weight increase, edge weight decreases, the mixture of them. For the edge updates belonging to different categories, various operations are processed. The main feature of these approaches is recognize the affected vertices, such as the intelligent semidynamic DSP algorithm named BallString in \cite{narvaez2001new}, the fully dynamic algorithm called DynamicSWSF-FP proposed in \cite{pallottinonew}.

However, the above algorithms have obvious disadvantages. When the scale of the network becomes very big or the weights of multiple edges decrease while that of other multiple edges increase, the procedure analyzing the affected vertices will become very complex. Secondly, from the practical viewpoint, this will cost lots of time. Especially in recent years, the networks with big scale become more and more. This may still cause long latency and unnecessary overheads. As a consequence, it is meaningful to explore new methods to handle the SPTs problem. Recently, a large amoeboid organism, the plasmodium of {\em Physarum polycephalum}, has been shown to be capable of solving many graph theoretical problems \cite{tero2010rules,Baumgarten,Watanabe2011,tero2006physarum}, including finding the shortest path \cite{nakagaki2007minimum,adamatzky2012slime,zhang2013solving,zhang2013route}, network design \cite{adamatzky2011brazilian,gunji2008minimal,adamatzky2011approximating}, population migration \cite{adamatzky2012world} and others \cite{aono2011amoeba,adamatzky2008growing,jones2010characteristics,aono2010amoeba,shirakawa2012multi}. Moreover, this organism has been shown to be able to form networks with features comparable to or better than the Tokyo rail network \cite{tero2010rules}. In addition, Baumgarten has proved the mass of mold will eventually convergence to the shortest path of the network that the mold
lies on \cite{Baumgarten}. Inspired by this intelligent organism, a path finding mathematical model has been established \cite{Tero2007}. To the best of knowledge, the amoeba model is not used to deal with SPTs by now.

In what follows, based on the amoeba model, an adaptive amoeba appraoch to SPTs in dynamic graphs is presented. The main characteristic of the propose method is its adaptivity. More specially, the algorithm can recognize the affected vertices and reconstruct them spontaneously. Those unaffected nodes will not be computed again. We will introduce how to implement amoeba model to deal with SPTs in the following sections.


\section*{Results}
\textbf{System Environment and Data Sets.} In order to evaluate the performance of the proposed method, we introduce the experimental environment and the problem instance generator is presented. Besides, the performance of our algorithm is compared with the Label Setting algorithm \cite{meyer2003average} and Bellman-Ford algorithm \cite{nguyen2007multicast}.

The proposed adaptive amoeba algorithm for shortest path tree in dynamic graphs is tested on networks with random and varying topologies through computer simulations using Matlab on an Intel Pentium Dual-Core E5700 processor (3.00 GHz) with 2 GB of RAM under Windows Seven. The random directed graphs can be generated using the {\em erdos.renyi.game} function of the igraph package in R language (for details, please refer to http://igraph.sourceforge.net/doc/R/erdos.renyi.game.html). The weight for an edge is randomly generated ranging from 1 to 1000. The data for random graph is shown in Table \ref{dataset}.

\textbf{Performance Indicators.} In each testing graph, the first node in the random generated graph is denoted as the source node. Then, a set $\lambda $ of edges is randomly selected to decrease or increase their edge weights. If both the increased case and the decreased case associated with the edge weights appear in the network, we denote $\lambda $ as mixed. In this paper, we pay attention to the CPU runtime for each algorithm. In order to examine the efficiency of the presented method, the following parameters are taken into consideration.

\begin{itemize}
  \item Graph size (\emph{graphsize}). It denotes the size of the network. We will focus on how the CPU runtime changes with the change of network size.
  \item Ratio of updated edges (\emph{rue}). This parameter represents the percentage of updated edges occupied in the whole network. For instance, in a network with 1000 edges, when 100 edges get their weight updated, then we will say \emph{rue} is 0.1.
  \item Ratio of changed weight (\emph{rcw}). This variable reflects the degree of the changed weight which is decreased or increased from its original value. As for this parameter, in the increased case, the \emph{rcw} for the edge weight ranges from 1 to a large number 10. For example, assume the original weight value associated with the edge is 100. If the parameter \emph{rcw} is 1, the updated weight for this edge will be 200. In the decreased case, the \emph{rcw} for the edge weight ranges from 0 to 0.9.
\end{itemize}

In order to evaluate the performance of the proposed method, we compare it with modified Dijkstra algorithm. We will show how the parameter \emph{rue} affects the investigated algorithms. For the following computational results, the program is run for 10 times for each instance.

\begin{figure*}[!ht]
\centering
\includegraphics[width=7in]{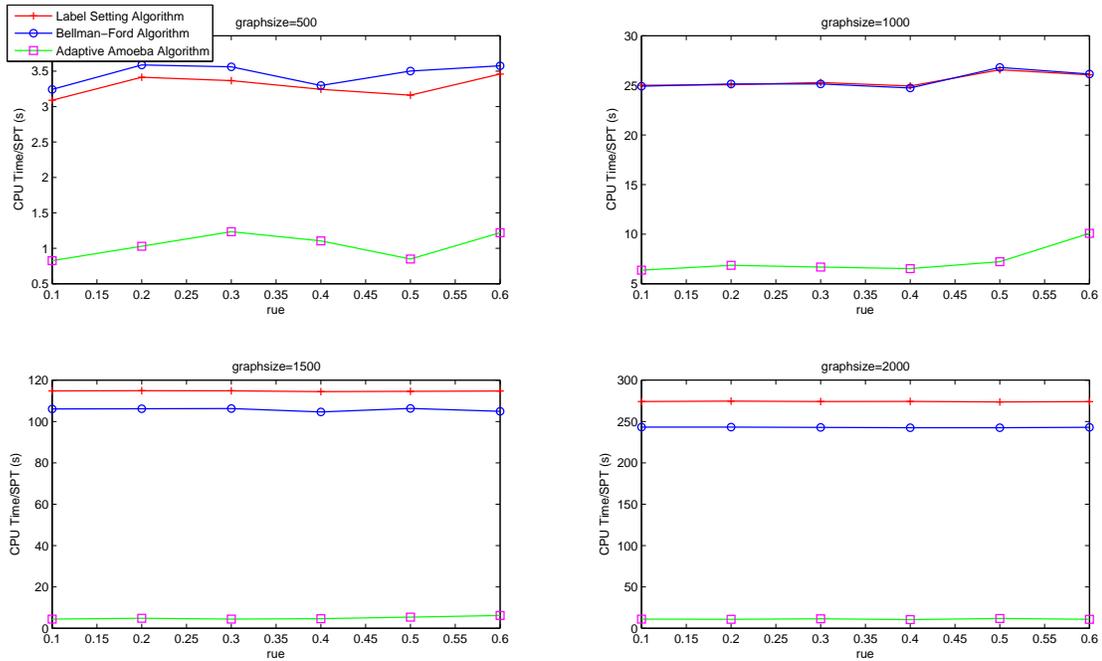}
\caption{Comparison in edge weight increases when the parameter \emph{rue} changes from 0.1 to 0.6 (here, the parameter \emph{rcw} is set a constant value $0.1$) }
\label{alphaIncrease}
\end{figure*}

\textbf{Edge Weight Increases.} For the increase case, Fig. \ref{alphaIncrease} shows the influence of \emph{rue} on the networks with different sizes when the parameter \emph{rue} changes from 0.1 to 0.6 (here, the parameter \emph{rcw} is set a constant value $0.1$). As can be seen in Fig. \ref{alphaIncrease}, in the increase cases, regardless of the change of the parameter \emph{rue}, the CPU time of all the three algorithms remain constant. On the other hand, considering CPU runtime, the Bellman-Ford algorithm has less computational time than the Label Setting algorithm in the four networks. The proposed adaptive amoeba algorithm outperforms when compared with the Label Setting algorithm and Bellman-Ford algorithm. The reason for this phenomenon is due to the change of the edge weights. When the weights associated with the edge change, the proposed adaptive amoeba algorithm can recognize the affected vertices and reconstruct them spontaneously. However, for the Label Setting Algorithm and Bellman-Ford Algorithm, it must reconstruct the whole network, which consumes more time. As a result, there is a big gap between the CPU runtime of the proposed method and that of Label Setting algorithm and Bellman-Ford algorithm.

\begin{figure*}[!ht]
\centering
\includegraphics[width=7in]{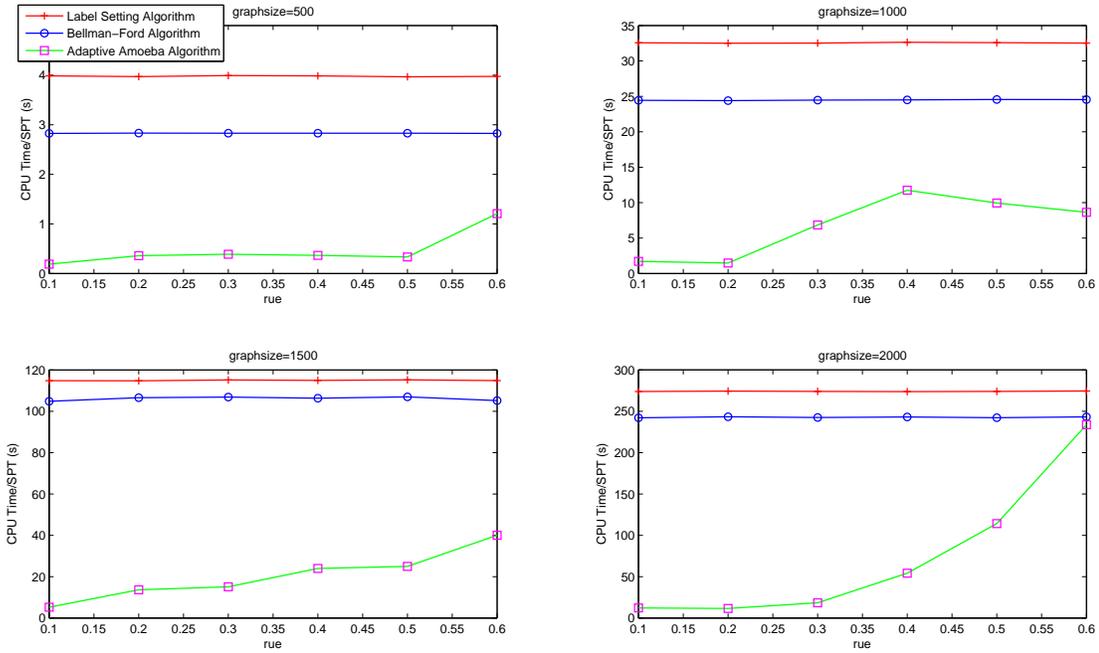}
\caption{Comparison in edge weight increases when the parameter \emph{rcw} changes from 0.1 to 0.6 (here, the parameter \emph{rue} is set a constant value $0.2$) }
\label{betaIncrease}
\end{figure*}

In what follows, we focus on how the change of the parameter \emph{rcw} affects the CPU runtime of the different algorithms in the increase case. As shown in Fig. \ref{betaIncrease}, it shows the influence of the parameter \emph{rcw} on the three algorithms' computational time when the parameter \emph{rcw} changes from 0.1 to 0.6 (here, the parameter \emph{rue} is set a constant value $0.2$). As we can see, the parameter \emph{rcw} has different influences on the three algorithms. For the Label Setting algorithm and Bellman-Ford algorithm, the computational time remains relatively constant, which is shown as straight lines in Fig. \ref{betaIncrease}. On the contrary, for the proposed adaptive amoeba algorithm, it is influenced strongly by the parameter \emph{rcw}, especially when the graph size is greater than 1500. From Fig. \ref{betaIncrease}, it can be seen that the CPU runtime of the presented method becomes more and more with the increase of parameter \emph{rcw}. The reason lies that when the parameter \emph{rcw} becomes bigger and bigger, the edge length is increased a lot from its original value. As a result, most of the paths in the SPT need to be recomputed. Although the CPU time of the proposed adaptive amoeba algorithm increases a lot, it is still less than that of the Label Setting algorithm and Bellman-Ford algorithm.

In summary, in the increase case, the change of the parameter \emph{rue} has little influence on these algorithms. The adaptive amoeba algorithm proposed in this paper has the least CPU runtime when dealing with the networks with different sizes. On the contrary, the change of the parameter \emph{rcw} has great influence on the CPU runtime of the proposed method. The CPU runtime of the presented method becomes bigger with the increase of parameter \emph{rcw}.

\begin{figure*}[!ht]
\centering
\includegraphics[width=7in]{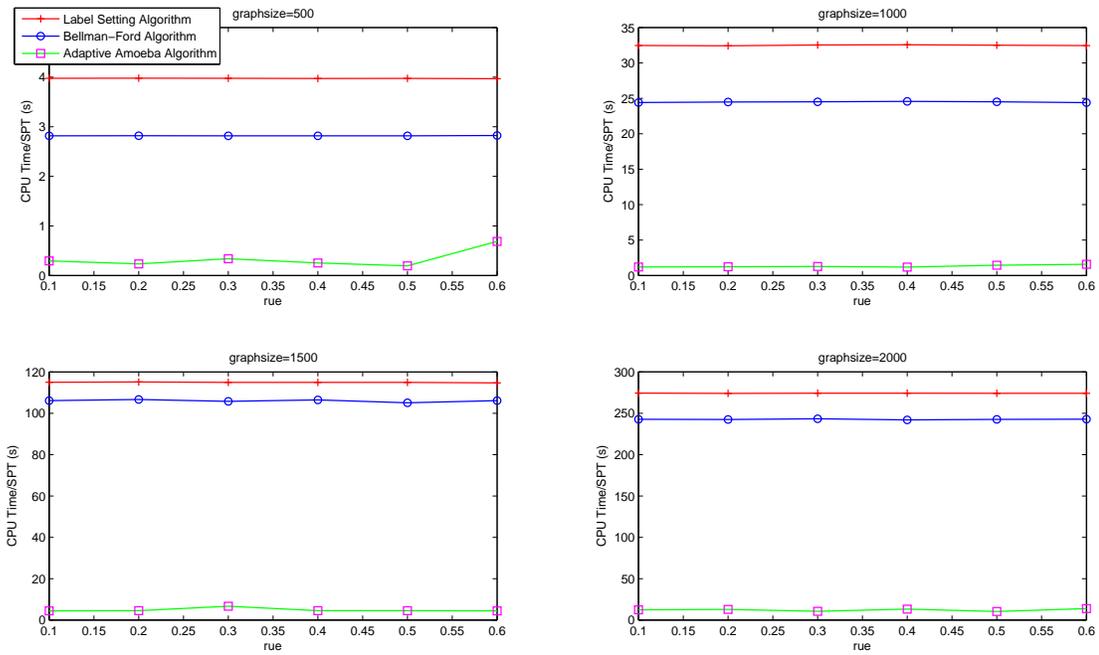}
\caption{Comparison in edge weight decreases when the parameter \emph{rue} changes from 0.1 to 0.6 (here, the parameter \emph{rcw} is set a constant value $0.1$) }
\label{alphaDecrease}
\end{figure*}

\begin{figure*}[!ht]
\centering
\includegraphics[width=7in]{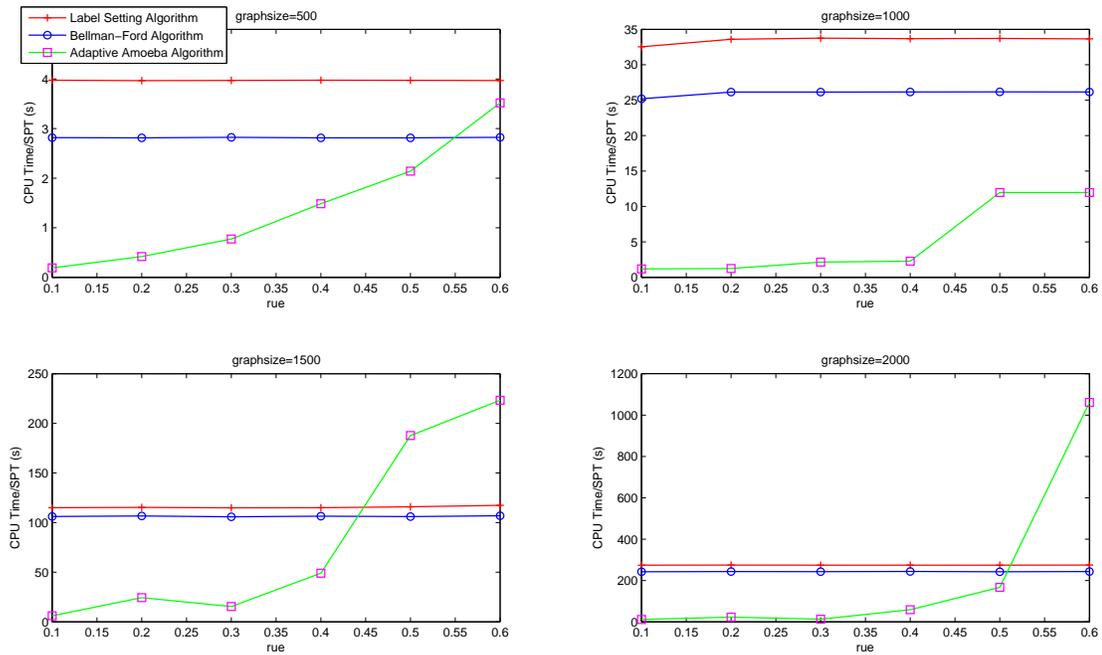}
\caption{Comparison in edge weight increases when the parameter \emph{rcw} changes from 0.1 to 0.6 (here, the parameter \emph{rue} is set a constant value $0.2$) }
\label{betaDecrease}
\end{figure*}

\textbf{Edge Weight Decreases.} Similarly, in the decrease case, we have observed the influence of the two parameters on the CPU runtime of the above three algorithms. Figs. \ref{alphaDecrease} and \ref{betaDecrease} show the experimental results, when the parameters \emph{rue}, \emph{rcw} change respectively.

As can be seen in Fig. \ref{alphaDecrease}, the parameters are set the same as that of the edge weight increases. From Fig. \ref{alphaDecrease}, it can be concluded that, in the decrease case, when the parameter \emph{rue} changes from 0.1 to 0.6, the CPU runtime of the three algorithms fluctuates slightly in the networks with different sizes. In other words, the parameter \emph{rcw} has very little effect on the computational efficiency of these algorithms. The proposed method has obvious advantage over the Label Setting algorithm and Bellman-Ford algorithm when dealing with the SPT in dynamic graphs.

As shown in Fig. \ref{betaDecrease}, when the parameter \emph{rcw} changes from 0.1 to 0.6, the CPU runtime of the proposed method fluctuates strongly. When \emph{rcw} increases from 0.1 to 0.6, the CPU runtime for the adaptive amoeba algorithm deteriorates rapidly. In the network with graph size equal to 500, 1500, 2000, when \emph{rcw} reaches 0.6, it can be observed that the performance of the Label Setting algorithm and Bellman-Ford algorithm outperform compared to the proposed method. It can be observed that when \emph{rcw} is less than a certain threshold value, the proposed method has better performance. The threshold value varies with the size of a graph. When \emph{rcw} is bigger than 0.4, it is more appropriate to adopt the Label Setting algorithm and Bellman-Ford algorithm. The reason why the performance of the proposed algorithm gets worse when \emph{rcw} is bigger than the threshold value is that more edges are influenced, and more time is spent to reconstruct the SPT and reallocate the flux associated with each edge in the amoeba algorithm.

In summary, in the decreased case, the situation is similar to that of the increased case. The parameter \emph{rcw} has different effect with that of parameter \emph{rue}. Besides, the performance of the adaptive amoeba algorithm gets worse when \emph{rcw} is more than a threshold value. As a consequence, it is appropriate to carry out different algorithms according to the specific value of this parameter.

\textbf{Mixed Edge Weight Changes.} In the mixed case, both the increased case and the decreased case share the same percentage in the total ratio of updated edges. Simply speaking, suppose there are 200 edges in the network whose weights change, then 100 of them increase their edge weights while the left 100 edges' weights decrease. As for the parameter \emph{rcw}, it is implemented for both the increased case and the decreased case. It is to say suppose \emph{rcw} is $0.1$, then half of the randomly selected edges will increase their weights by 10 percentages while the left edges will decrease their weights by 10 percentages.

\begin{figure*}[!ht]
\centering
\includegraphics[width=7in]{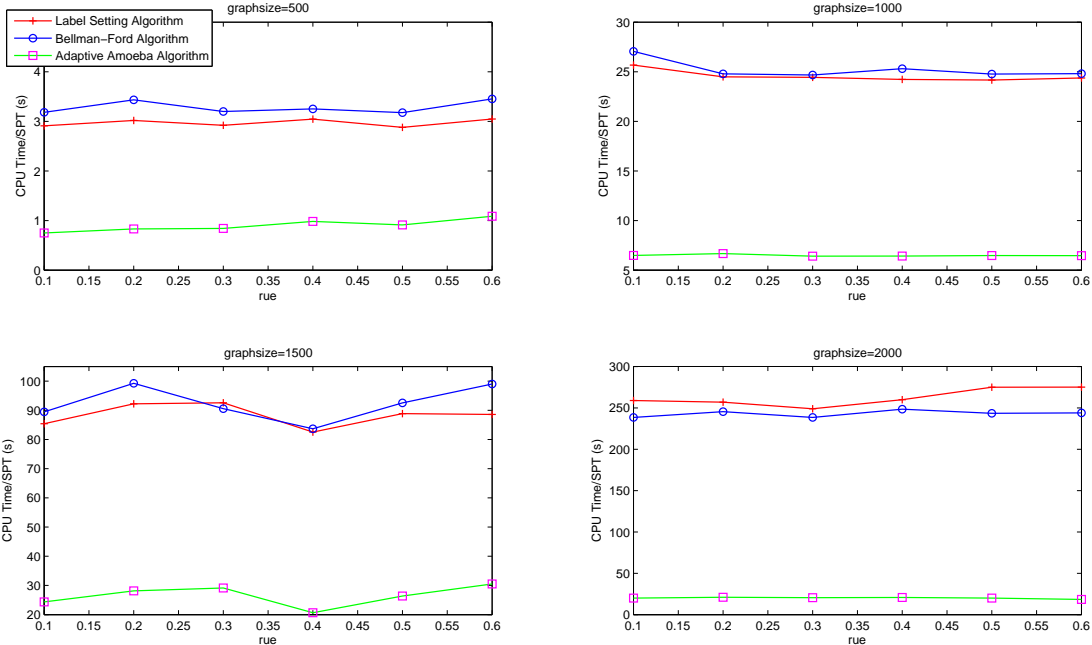}
\caption{Comparison in mixed change of the edge weight when the parameter \emph{rue} changes from 0.1 to 0.6 (here, the parameter \emph{rcw} is set a constant value $0.1$) }
\label{alphaMixture}
\end{figure*}

Fig. \ref{alphaMixture} shows the CPU runtime of the three algorithms when the parameter \emph{rue} changes from 0.1 to 0.6. Similarly, it has slight influence like in the increased and decreased cases. Obviously, the adaptive amoeba algorithm outperforms when compared with the other algorithms.

\begin{figure*}[!ht]
\centering
\includegraphics[width=7in]{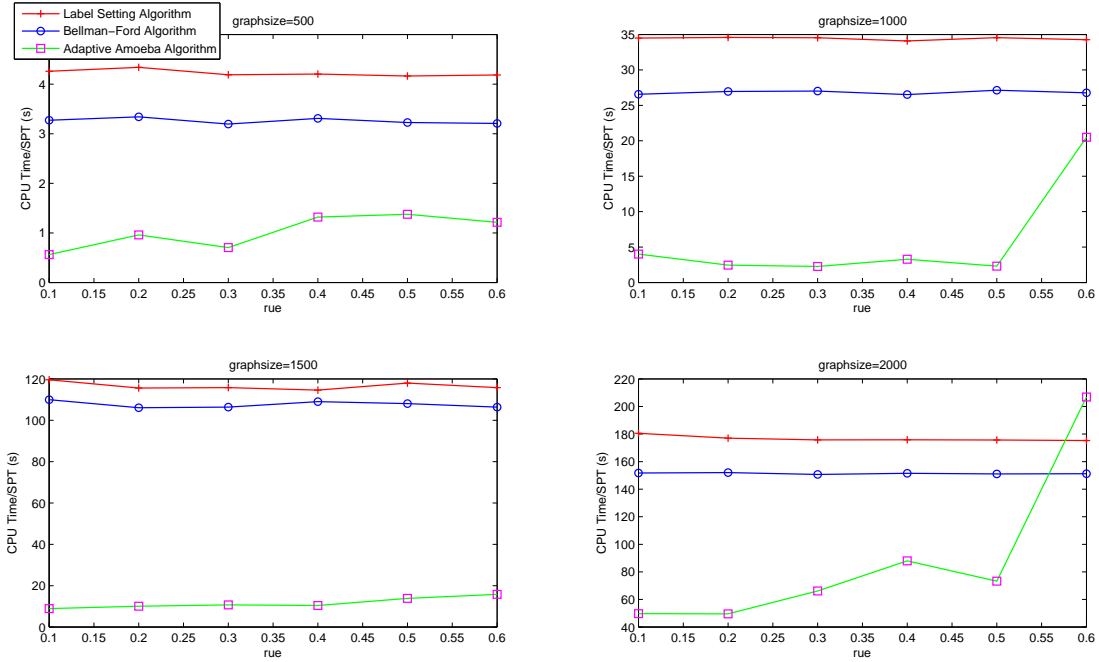}
\caption{Comparison in mixed change of the edge weight when the parameter \emph{rcw} changes from 0.1 to 0.6 (here, the parameter \emph{rue} is set a constant value $0.2$) }
\label{betaMixture}
\end{figure*}

As for the change of parameter \emph{rcw}, Fig. \ref{betaMixture} displays the computational results of the three algorithms. It can be concluded that the Label Setting algorithm and Bellman-Ford algorithm remain constant, regardless of the changed weights. For the adaptive amoeba algorithm, it fluctuates greatly when \emph{rcw} changes. As can be seen in Fig. \ref{betaMixture}, when the \emph{rcw} is less than 0.5, the proposed method outperforms certainly than the other two algorithms. When \emph{rcw} reaches 0.6 and the graph size reaches 2000, the CPU runtime of the presented algorithm has spent more time than the other two algorithms. In the four sub figures, the results are not consistent. It is due to the difference of the network topology. Every time, the affected edges are randomly selected, which may lead to different influences on the network. In one word, similar to the increased and decreased case, the parameter \emph{rue} affect the CPU runtime a little while \emph{rcw} playing an important role regarding the CPU runtime.

\section*{Discussion}
In this paper, we investigated the previous algorithms used to deal with SPT problem in dynamic graphs. These algorithms solving SPT problem with edge updates by identifying the affected nodes and reconstruct the shortest path among these nodes. However, when these algorithms are faced with the network with big scale, on the one hand, this procedure becomes very complicated. On the other hand, it cost much time. In order to address the above problems, we proposed a fully adaptive amoeba algorithm for solving SPT in dynamic graphs. The efficiency of the presented algorithm is demonstrated by implementing it in all the edge updates including the increase, decrease, mixed change of the edge weight. In order to evaluate the performance of the proposed method, we conducted experiments on randomly generated graphs, in terms of the CPU execution time. Moreover, we compared our algorithm with the existing algorithms, such as the Label Setting algorithm, Bellman-Ford algorithm. The purpose of the experiment is to obverse how these algorithms behave for different graph sizes and various mixes of changed edges. We randomly generate
four graphs with different sizes: $500, 1000, 1500, 2000$.

We evaluate the performance of the three algorithms according to three factors: \emph{graphsize}, ratio of updated edges (\emph{rue}), ratio of changed weight (\emph{rcw}). For the increased case, the change of the parameter \emph{rue} has little influence on these algorithms while the change of the parameter \emph{rcw} has great influence on the CPU runtime of the proposed method. For the decreased case, the influence of parameter \emph{rue} is similar to that of the increased case. On the other hand, the performance of the adaptive amoeba algorithm gets worse when \emph{rcw} is more than a threshold value. As for the other left algorithms, they are slightly influenced by parameter \emph{rcw}. For the mixed change of the edge weights, the influences of parameters \emph{rue} and \emph{rcw} on the computational time are consistent with that of the increased and decreased cases. We conclude the following for the above three algorithms. For the increased case, decreased case, and mixed change case, in spite of the ratio of updated edges, the proposed method has the best overall performance. On the contrary, when the parameter \emph{rcw} changes,  it is appropriate to carry out different algorithms according to the specific value of \emph{rcw}.

\section*{Methods}

\textbf{\textit{Physarum Polycephalum} Inspired Shortest Path Finding Model.}\label{Section2} {\em Physarum Polycephalum} is a large, single-celled amoeboid organism forming a dynamic tubular network connecting the discovered food
sources during foraging. The mechanism of tube formation can be
described as: tubes thicken in a given direction when shuttle
streaming of the protoplasm persists in that direction for a
certain time. It implies positive feedback between flux and tube
thickness, as the conductance of the sol is greater in a thicker
channel. With this mechanism, a mathematical model illustrating the shortest path finding has been constructed \cite{Tero2007}.

Suppose the shape of the network formed by the \textit{Physarum} is represented by a graph, in
which a plasmodial tube refers to an edge of the graph and a junction
between tubes refers to a node. Two special nodes labeled as $N_1$, $N_2$ act as the starting node and ending node respectively. The other nodes are labeled as $N_3,N_4,N_5,N_6$ etc. The edge between node $N_i$ and $N_j$ is expressed as $M_{ij}$. The parameter $Q_{ij}$ denotes the flux through tube $M_{ij}$ from node $N_i$ to $N_j$. Regard the flow along the tube as an approximately
poiseuille flow, the flux ${Q_{ij}}$ can be expressed as:

\begin{equation}\label{flux}
{Q_{ij}} = \frac{{{D_{ij}}}}{{{L_{ij}}}}({p_i} - {p_j})
\end{equation}
where ${p_i}$ is the pressure at the node ${N_i}$, ${D_{ij}}$
is the conductivity of the tube ${M_{ij}}$, $L_{ij}$ is its length.

By considering that the inflow and outflow must be balanced, we have:
\begin{equation}\label{sum}
\sum {{Q_{ij}}}  = 0(j \ne 1,2)
\end{equation}

  For the source node ${N_1}$ and the sink node ${N_2}$ the
following two equations hold
\begin{equation}\label{soucesink}
\sum\limits_i {{Q_{i1}}}  + {I_0} = 0
\end{equation}

\begin{equation}\label{soucesink1}
\sum\limits_i {{Q_{i2}}} -{I_0} = 0
\end{equation}
where ${I_0}$ is the flux flowing from the source node and ${I_0}$ is a
constant value here.

  In order to describe such an adaptation of tubular thickness we
assume that the conductivity ${D_{ij}}$ changes over time
according to the flux ${Q_{ij}}$. The following equation for the
evolution of ${D_{ij}(t)}$ can be used
\begin{equation}\label{changeovertime}
\frac{d}{{dt}}{D_{ij}} = f(|{Q_{ij}}|) - r{D_{ij}}
\end{equation}
where $r$ is a decay rate of the tube. It can be obtained that the
equation implies that the conductivity ends to vanish if there is
no flux along the edge, while it is enhanced by the flux. The $f$
is monotonically increasing continuous function satisfying $f(0) =
0$.

  Then the network poisson equation for the pressure can be
obtained from the Eq. (\ref{flux}-\ref{soucesink1}) as follows:
\begin{equation}\label{getresult}
\sum\limits_i {\frac{{{D_{ij}}}}{{{L_{ij}}}}({p_i} - {p_j})}  =
\left\{ {\begin{array}{*{20}{c}}
   { + 1} & {for} & {j = 1,}  \\
   { - 1} & {for} & {j = 2,}  \\
   0 & {otherwise} & {}  \\
\end{array}} \right.
\end{equation}

  By setting ${p_2}$=0 as a basic pressure level, all ${p_i}$
can be determined by solving Eq. (\ref{getresult}) and ${Q_{ij}}$
can also be obtained.

  In this paper, $f(Q) = |Q|$ is used. With the flux calculated, the conductivity can be derived,
where Eq. (\ref{eql6}) is used instead of Eq.
(\ref{changeovertime}), adopting the functional form $f(Q) = |Q|$.
\begin{equation}\label{eql6}
\frac{{D_{ij}^{n + 1} - D_{ij}^n}}{{\delta t}} = |Q| - D_{ij}^{n +
1}
\end{equation}

The amoeba model is modified to solve SPTs in dynamic graphs. The algorithm is composed of three parts. First of all, the amoeba model is extended to solve the shortest path in the directed networks. Then a modified model is presented to handle SPTs in static directed networks. Finally, we analyze the changing trend of the amoeba algorithm is applied to deal with SPTs in dynamic graphs.

\textbf{Shortest Path Problem in Directed Network.} It is observed that the original amoeba model can only handle the shortest path problem (SPP) in the undirected network according to \cite{tero2006physarum,nakagaki2007minimum,Tero2007}. As a result, we need to extend its application field to directed networks by modifying the original model.

Let \(G = \left(N, E, L \right)\) be a directed network, where $N$ denotes a set of $n$ nodes, $E$ denotes an edge set with $m$ directed edges, and $L$ denotes a weight set for $E$. Given a source node $s$ and a sink node $t$, the directed shortest path problem can be defined as how to find a path from $s$ to $t$, which only consists of directed edges of $E$, with the minimum sum of weights on the edges.

The original amoeba model is designed to solve the shortest path problem in the undirected graphs. For the directed graphs, the following equations are constructed according to the Kirchhoff's laws:
\begin{equation}\label{newFlow}
\sum\limits_{j \in N} {\left( {\frac{{{D_{ij}}}}{{{L_{ij}}}} + \frac{{{D_{ji}}}}{{{L_{ji}}}}} \right)\left( {{p_i} - {p_j}} \right)}  = \left\{ \begin{array}{l}
  + 1\;\;\;\;\;for\;i = s \\
  - 1\;\;\;\;\;for\;i = t \\
 0\;\;\;\;\;\;\;otherwise \\
 \end{array} \right.
\end{equation}
where $s$ denotes the starting node, $t$ denotes the ending node, $L_{ij}$ is the length of the edge ${M_{ij}}$, $D_{ij}$ is the conductivity of the edge $M_{ij}$. In the directed graphs, $M_{ij}$ is different from that of the undirected networks. Here, $M_{ij}$ denotes the tube starting from node $i$ to node $j$. As a result, the way that $D_{ij}$ is initialized in the directed graphs is different from that in the undirected networks. If ${M_{ij}} \in E$, then \({D_{ij}} = 1\). Otherwise, \({D_{ij}} = 0\).

In what follows, the flux $Q_{ij}$ of every edge can be obtained. The direction of the edge is related with the pressure of each node. Naturally, the flux starts at the node with high pressure and ends at the node with low pressure. In the directed graphs, in order to keep the direction of each edge, the following check procedure is inevitable. Assume there is an edge $M_{ij}$ starting from node $i$ and ending at node $j$. If the pressure $p_j$ is larger than $p_i$, it means the flux is flowing from node $j$ to node $i$, which is opposite with the direction of the edge in the directed graph. Once the phenomenon is found, then the flux needs to be cut off. As a consequence, we change the flux of this edge $Q_{ij}$ to be 0. Next, the following Eq. (\ref{newConductivity}) for the evolution of $D_{ij}(t)$ can be summarized. As can be seen, the changing process of $D$ is continuous. In the process of the algorithm, a discrete procedure is applied as shown in Eq. (\ref{computerConductivity}). The general flow of the proposed method is detailed in Table \ref{algorithm1}.

\begin{equation}\label{newConductivity}
\frac{{d{D_{ij}}\left( t \right)}}{{dt}} = \left\{ \begin{array}{l}
 {Q_{ij}}\left( t \right) - {D_{ij}}\left( t \right)\;\;\;\;\;{p_i}\left( t \right) \ge {p_j}\left( t \right) \\
  - {D_{ij}}\left( t \right)\;\;\;\;\;\;\;\;\;\;\;\;\;\;{p_i}\left( t \right) < {p_j}\left( t \right) \\
 \end{array} \right.
\end{equation}

\begin{figure*}[ht]
\begin{equation}\label{computerConductivity}
{D_{ij}}\left( {n + 1} \right) = \left\{ \begin{array}{l}
 \left( {{Q_{ij}}\left( n \right) - {D_{ij}}\left( n \right)} \right)\;*\Delta t + {D_{ij}}\left( n \right)\;\;\;\;{p_i}\left( n \right) \ge {p_j}\left( n \right) \\
 \left( {0 - {D_{ij}}\left( n \right)} \right)\;*\Delta t + {D_{ij}}\left( n \right)\;\;\;\;\;\;\;\;\;\;\;{p_i}\left( n \right) < {p_j}\left( n \right) \\
 \end{array} \right.
\end{equation}
\end{figure*}
where $\Delta t$ is the time interval, $0 < \Delta t < 1$.

There are several possible solutions to decide when to stop execution of Algorithm 1, such as the maximum number of iterations is arrived, conductivity of each tube converges to 0 or 1, flux through each tube remains unchanged, etc. In this paper, when the conductivity matrix changes between the current iteration and the previous iteration very little (in other words, $\sum\limits_{i = 1}^N {\sum\limits_{j = 1}^N {\left| {D_{ij}^{n } - D_{ij}^{n-1}} \right|} }  \le \delta$, where $\delta $ is a threshold), then the program will stop.

\textbf{Proofs of {\em Physarum Polycephalum} Model in Directed Networks.} In undirected graphs, the model is able to converge to an equilibrium state and Baumgarten \cite{Baumgarten} has proved its stability . In directed graphs, a lot of our experiments have shown that, it will also convergence to an equilibrium state in directed networks and the process of the convergence is similar to that of the undirected graphs. In this section, we prove that all the flux flowing from $s$ to $t$ consists of the shortest path in the directed networks when the network reaches the equilibrium state.

First of all,  the following parameters are defined:

\(F\left( t \right)\): \(F\left( t \right)\) is one set of edges whose $Q_{ij}$ is non-negative number at time $t$. Let \(F = \mathop {\lim }\limits_{t \to \infty } F\left( t \right)\).

\(B\left( t \right)\): \(B\left( t \right)\) is another set of edges whose $Q_{ij}$ is is negative number at time $t$. Let \(B = \mathop {\lim }\limits_{t \to \infty } B\left( t \right)\).

\({u_{ij}}\left( t \right)\): \({u_{ij}}\left( t \right)\) is the pressure difference between node $i$ and node $j$ at time $t$. \({u_{ij}}\left( t \right) = {p_i}\left( t \right) - {p_j}\left( t \right)\). Let \({u_{ij}} = \mathop {\lim }\limits_{t \to \infty } {u_{ij}}\left( t \right)\).

\textbf{lemma 3.1.} For any directed edge $M_{ij}$, \(\mathop {\lim }\limits_{t \to \infty } \left( {{Q_{ij}}\left( t \right) - {D_{ij}}\left( t \right)} \right) = 0\).

\textbf{\textit{Proof}} When $t$ goes to a infinite number, the network reaches stable. As a result, for any $M_{ij}$, \(\mathop {\lim }\limits_{t \to \infty } \frac{{d{D_{ij}}\left( t \right)}}{{dt}} = 0\). According to Eq. (\ref{newConductivity}), it can be seen that:
\[\left\{ \begin{array}{l}
\mathop {\lim }\limits_{t \to \infty } \left( {{Q_{ij}}\left( t \right) - {D_{ij}}\left( t \right)} \right) = 0\;\;\;\;\;M_{ij} \in F\\
\mathop {\lim }\limits_{t \to \infty } {D_{ij}}\left( t \right) = 0\;\;\;\;\;\;\;\;\;\;\;\;\;\;\;\;\;\;\;M_{ij} \in B\;
\end{array} \right.\]

When \(j \in B\), \(\mathop {\lim }\limits_{t \to \infty } {Q_{ij}}\left( t \right) = \mathop {\lim }\limits_{t \to \infty } \frac{{{D_{ij}}\left( t \right){u_{ij}}\left( t \right)}}{{{L_{ij}}}} = 0\).

Lemma 3.1 is established.

\textbf{lemma 3.2.} For any $M_{ij}$, if \(\mathop {\lim }\limits_{t \to \infty } {Q_{ij}}\left( t \right) \ne 0\), then \({M_{ij}} \in F\) and \({u_e} = {L_e}\).

\textbf{\textit{Proof}} According to lemma 3.1, if \(\mathop {\lim }\limits_{t \to \infty } {Q_{ij}}\left( t \right) \ne 0\), then \(\mathop {\lim }\limits_{t \to \infty } {D_{ij}}\left( t \right) = \mathop {\lim }\limits_{t \to \infty } {Q_{ij}}\left( t \right) \ne 0\).

Based on Eq. (\ref{flux}), it can be seen that \({u_{ij}} = {L_{ij}} > 0\) and \({M_{ij}} \in F\).

Lemma 3.2 is established.

\textbf{lemma 3.3.} If \({M_{ij}} \in F\), then \({u_{ij}} \le {L_{ij}}\).

\textbf{\textit{Proof}} According to lemma 3.2, when \(\mathop {\lim }\limits_{t \to \infty } {Q_{ij}}\left( t \right) \ne 0\), lemma 3.3 is established.

 When \(\mathop {\lim }\limits_{t \to \infty } {Q_{ij}}\left( t \right) = 0\) and \({M_{ij}} \in F\), then \(\mathop {\lim }\limits_{t \to \infty } {D_{ij}}\left( t \right) = 0\), $\exists T,\forall t > T,\frac{{d{D_{ij}}\left( t \right)}}{{dt}} < 0$. Thus, \(M_{ij} \in F\left( t \right)\). According to Eq. (\ref{newConductivity}), we can obtain \({Q_{ij}}\left( t \right) < {D_{ij}}\left( t \right)\).

According to Eq. (\ref{flux}), it can be seen \({u_{ij}}\left( t \right) < {L_{ij}}\).

In summary, \({u_{ij}} \le {L_{ij}}\). Therefore, lemma 3.3 is established.

\textbf{lemma 3.4.}
When the model reaches the equilibrium state:

\begin{enumerate}[(i)]
\item  All the flow converges to some paths from $s$ to $t$.
\item  These directed paths have the same path length and their length is equal to $u_{st}$.
\item  These directed paths have the shortest path length.
\end{enumerate}

\textbf{\textit{Proof}}

\begin{enumerate}[(i)]
\item According to lemma 3.1 and lemma 3.2, in the equilibrium state, the directions of all the edges are in accordance with the flow directions. Consequently, the flow converges to the paths existing in the directed graph $G$.
\item Assume $v$ is one directed path which the flow converges to, $L_v$ is the length of $v$. According to lemma 3.2, there are \({u_{st}} = \sum\limits_{{M_{ij}} \in v} {{u_{ij}}}  = \sum\limits_{{M_{ij}} \in v} {{L_{ij}}}  = {L_v}\).
\item Assume $v$ is one path from $s$ to $t$, it is known to us that \({u_{st}} = \sum\limits_{{M_{ij}} \in v} {{u_{ij}}} \). For any edge $M_{ij}$, if \({u_{ij}} \ge 0\), then \({M_{ij}} \in F\). According to lemma 3.3, \({u_{ij}} \le {L_{ij}}\). If \({u_{ij}} < 0\), it is obviously seen that \({u_{ij}} \le {L_{ij}}\). Consequently, \({u_{st}} = \sum\limits_{{M_{ij}} \in v} {{u_{ij}}}  \le \sum\limits_{{M_{ij}} \in v} {{L_{ij}}}  = {L_v}\). It means that all the other paths' length is not less than $u_{st}$. Incorporating with (ii), these paths are the shortest ones.
\end{enumerate}

Based on the above proofs, lemma 3.4 is established.

\textbf{Shortest Path Tree Problem in Directed Networks.} The above algorithm can only compute the shortest path between two nodes at a time. Assume there are $N$ nodes in the network, if we want to construct the shortest path tree, the algorithm must be run for $N-1$ times. However, this will consume lots of time. In this section, after we modify the above model further, the shortest path tree can be constructed by running the algorithm one time. In what follows, the modified model used to find the shortest path tree is introduced.

In the original amoeba model, there are only one starting node $s$ and one ending node $t$ in it. If there are one starting node $s$ and all the other nodes are the ending nodes, the Kirchhoff¡¯s laws can be transformed as below:

\begin{equation}\label{SPTFlow}
\sum\limits_{i} {\left( {\frac{{{D_{ij}}}}{{{L_{ij}}}} + \frac{{{D_{ji}}}}{{{L_{ji}}}}} \right)} \left( {{p_i} - {p_j}} \right) = \left\{ \begin{array}{l}
  + 1\quad \;for\;j = s \\
 \frac{-1}{{N - 1}}\;\;\;for\;j \ne s\quad  \\
 \end{array} \right.
\end{equation}

The general flow of the modified model used to construct SPTs rooted at node $s$ are shown in Table \ref{algorithm2}. By this way, SPT can be constructed through running the program once.

%
%
%
%
%
%

\begin{figure*}[!ht]
\centering
\includegraphics[width=3in,angle=-90]{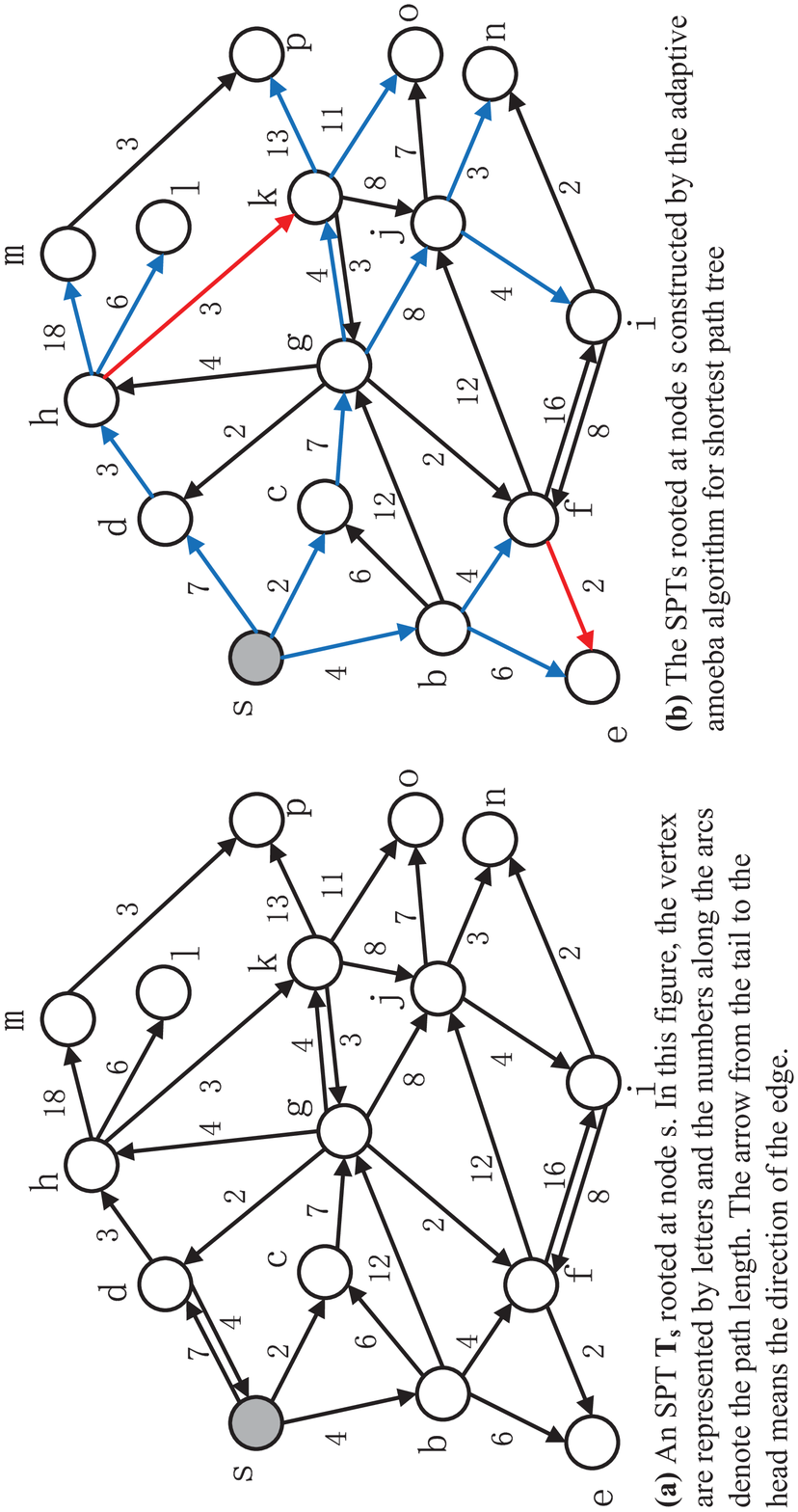}
\caption{The amoeba algorithm for SPT }
\label{original}
\end{figure*}

%

\begin{example}
In order to illustrate the algorithm shown in Table \ref{algorithm2}, an example is shown in Fig. \ref{original}(a). There are 16 nodes in this network. In this example, the amoeba model is applied to construct SPT rooted at node $s$. According to Algorithm 2, first of all, initialize the parameters, such as the length matrix $L$, the initial conductivity matrix $D$, the initial pressure matrix $Q$ etc. Then, according to Eq. (\ref{SPTFlow}), the pressure of each node can be obtained during the first iteration. In turn, the flux of each node can be computed according to Eq. (\ref{flux}). Next, the conductivity matrix during the following iteration can be constructed based on Eq. (\ref{eql6}). This procedure will continue until the flux of each arc do not change any more. Fig. \ref{flux} shows the flux variation of each edge during different iterations in the graph shown in Fig. \ref{original}(a) using the adaptive amoeba model. As we can see, the flux of some edges converges to $0$ during the iterations. Those edges whose flux are not equal to zero constitutes the following graph shown in Fig. \ref{original}(b).

\begin{figure}[!ht]
\centering
\includegraphics[width=3.5in]{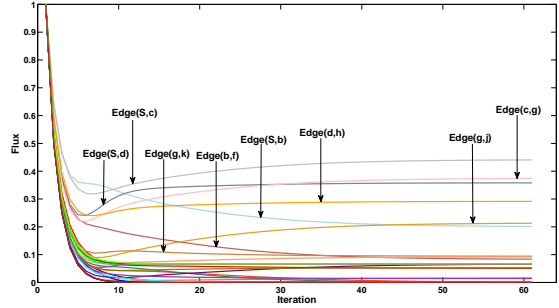}
\caption{Flux variation during different iterations in the graph shown in Fig. \ref{original}(a) using the adaptive amoeba model.}
\label{flux}
\end{figure}


Fig. \ref{original}(b) shows the SPTs rooted at vertex $s$. The trees represented by the edges in blue color are consistent with the result in \cite{chan2009shortest}. The edges with red color appear in the adaptive amoeba algorithm, but not in \cite{chan2009shortest}. The reason lies that there are more than one shortest paths between the rooted node $s$ and nodes $e$, $k$. For example, the length of the path $S \to b \to e$ is equal to that of the path $S \to b \to f \to e$. Similarly, the length of path $S \to d \to h \to k$ is equal to that of $S \to c \to g \to k$. This is the first advantage of our presented method.
\end{example}

The edge weight change can be divided into three categories: increase only, decrease only, the mixture of them. In this section, the adaptive amoeba algorithm for SPTs in dynamic graphs will be analyzed from the three perspectives.

\begin{figure*}[!ht]
\centering
\includegraphics[width=3in,angle=-90]{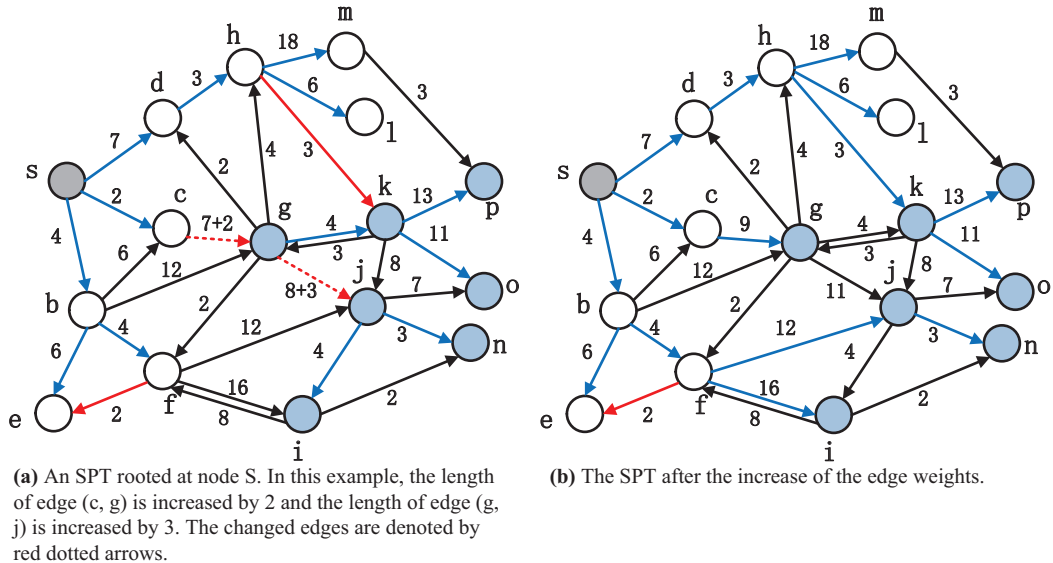}
\caption{The SPT with edge weight increase}
\label{increase}
\end{figure*}

\textbf{Edge Weight Increase.} As shown in Fig. \ref{increase}(a), the weights of edges $(c,g)$ and $(g,j)$ are increased. For traditional approaches to SPT in dynamic graphs, such as \emph{DynDijkInc, MBallString} \cite{chan2009shortest} etc, firstly, these methods locates all locally affected vertices and reconstruct the shortest path between these affected nodes. In this example, the vertices ${g, k, o, p, i, n}$ are affected, which are shown in blue circles in Fig. \ref{increase}(a). For the above methods, when the scale of the network become very big, on the one hand, the process to locate the affected nodes become very complex. On the other hand, it costs much time. For the adaptive amoeba algorithm proposed in this paper, the following Fig. \ref{FluxAfterIncrease} displays the changing trend of the flux existing in each edge.


\begin{figure}[!ht]
\centering
\includegraphics[width=3.5in]{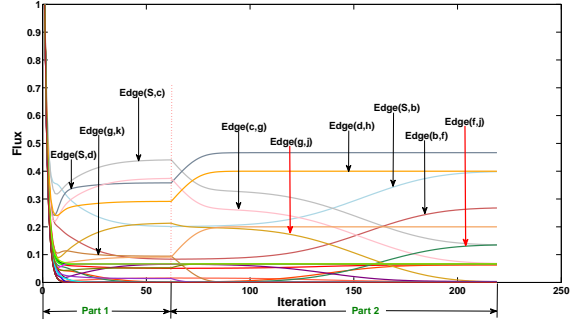}
\caption{The changing trend of the flux when the length of edge $(c,g)$ is increased by $2$ and the length of edge $(g,j)$ is increased by $3$}
\label{FluxAfterIncrease}
\end{figure}

As can be seen in Fig. \ref{FluxAfterIncrease}, the red doted line divides the whole process into two parts. The first part is the changing trend of the flux associated with each edge before the edge weights change. The second part illustrates how the flux associated with every edge change after the increase of edge weights. During the iteration process, the flux of the edge $(g,j)$ decreases to 0 while that of the edge $(f,j)$ become one of the edges constructing SPT. This example shows the adaptivity of the presented algorithm. It can recognize the affected vertices and reconstruct them spontaneously after the increase of the edge weights. Fig. \ref{increase}(b) displays the SPT after the increase of the edge weights and the result is consistent with that of \cite{chan2009shortest}.

\begin{figure*}[!ht]
\centering
\includegraphics[width=3in,angle=-90]{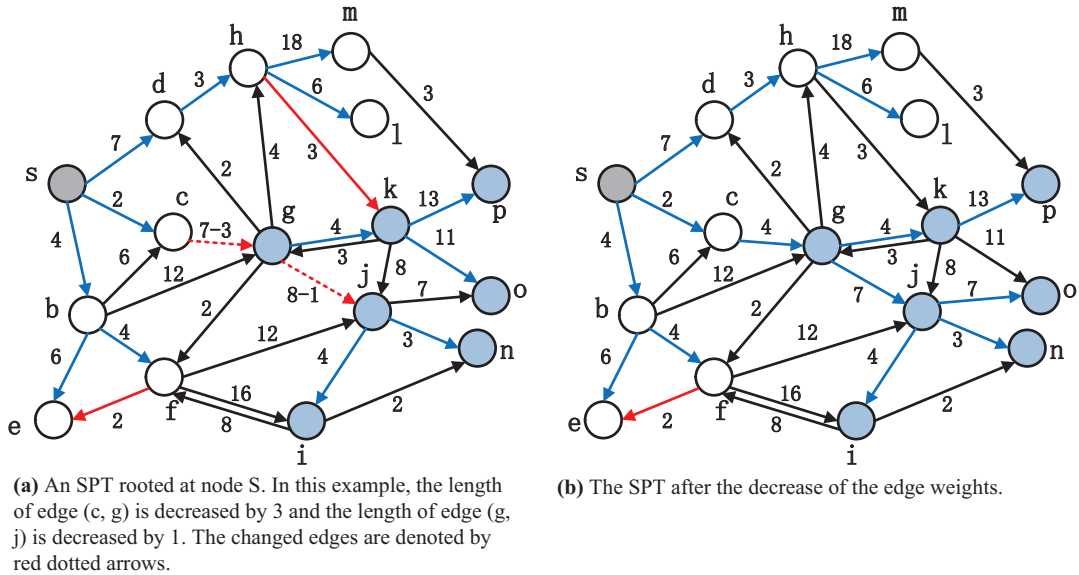}
\caption{The SPT with edge weight decrease}
\label{decrease}
\end{figure*}

\textbf{Edge Weight Decrease.} Given a graph shown in Fig. \ref{decrease}(a), there are a source vertex $S$, an SPT rooted at node $S$. The weights of edge $(c,g)$ and edge $(g,j)$ are decreased by $3$, $1$ respectively. Different from the increase case, the locally affected nodes can be predicted. In the decrease case, the traditional algorithms such as DynDijkstra, MBallString recognize the all affected heads and then recompute the shortest path among the affected heads. When the network becomes very big, the traditional algorithms are faced with the same problems mentioned in the increase case. For the adaptive amoeba algorithm, the following Fig. \ref{FluxAfterDecrease} illustrates the changing trend of the flux of each edge when the edge weights decrease.


\begin{figure}[!ht]
\centering
\includegraphics[width=3.5in]{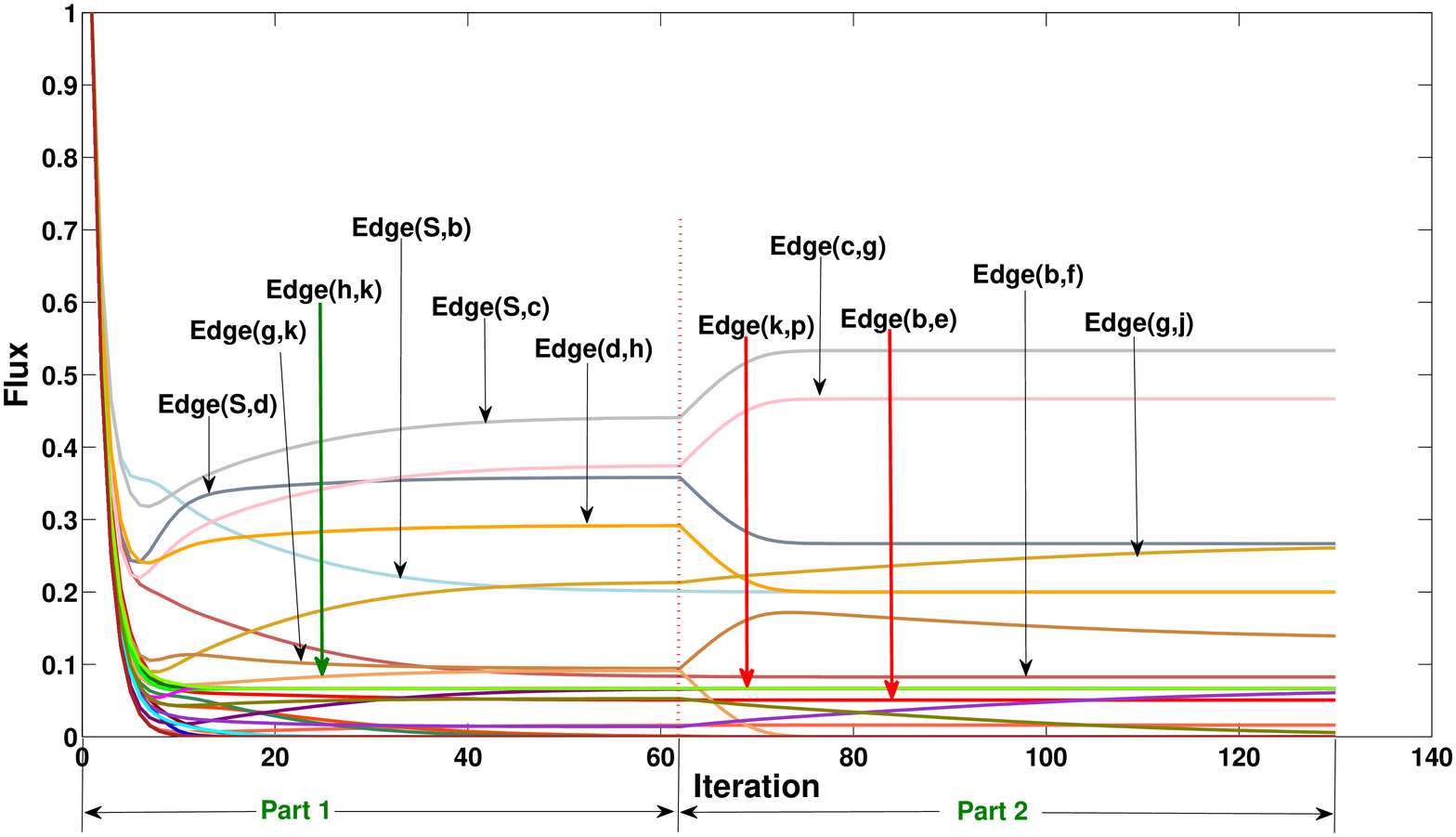}
\caption{The changing trend of the flux when the length of edge $(c,g)$ is decreased by $3$ and the length of edge $(g,j)$ is decreased by $1$}
\label{FluxAfterDecrease}
\end{figure}

As can be seen in Fig. \ref{FluxAfterDecrease}, the whole process can be separated into two procedures by the red dotted line: the part before edge weight decrease, the part after edge weight decrease. It can be seen that the flux associated with every edge changes when the edge weights decrease. All the edges can be divided into 2 categories: the edges which is not influenced by the edge weights decrease, such as edges $(k,p)$ and $(b,e)$ shown in Fig. \ref{FluxAfterDecrease} belongs to the first category; the second category includes the edges that are influence by the edge weights decrease, such as edges $(h,k)$, $(S,d)$, $(S,c)$ etc. From Fig. \ref{FluxAfterDecrease}, it can be concluded that the adaptive amoeba algorithm have the advantage of recognizing all the affected vertices spontaneously over traditional algorithms such as DynDijkstra, MBallString. Fig. \ref{decrease}(b) displays SPT after the edge weights decrease and the result is the same as that of \cite{chan2009shortest}.

\begin{figure*}[!ht]
\centering
\includegraphics[width=3in,angle=-90]{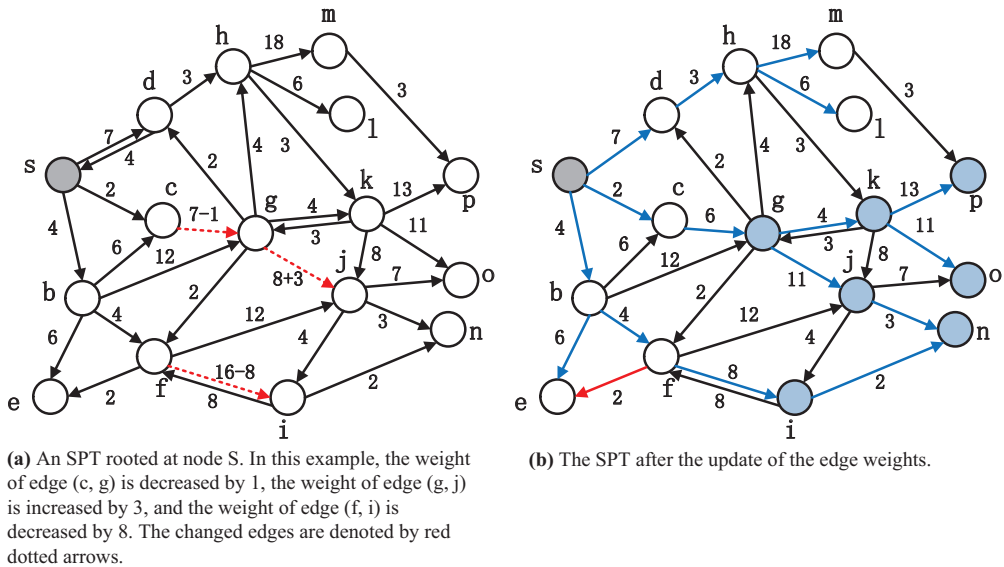}
\caption{The SPT with mixed edge weight changes}
\label{mixture}
\end{figure*}

\textbf{Mixed Edge Weight Changes.} In Fig. \ref{mixture}(a), the weights of the following edges updates: edge $(c,g)$ is decreased by $1$, the weight of edge $(g,j)$ is increased by $3$, and the weight of edge $(f,i)$ is decreased by 8. Based on this example, we will pay attention to the changing trend of the flux associated with each edge in the adaptive amoeba algorithm when both the decrease and increase of the edges appear in the graph.


\begin{figure}[!ht]
\centering
\includegraphics[width=3.5in]{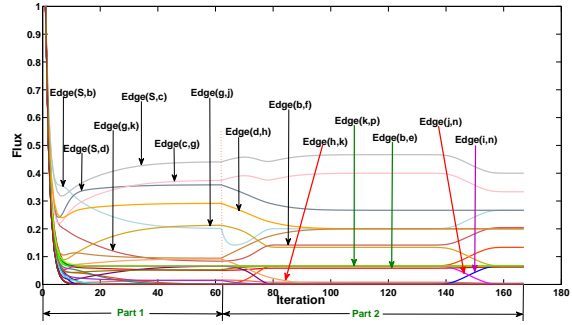}
\caption{The changing trend of the flux associated with each edge when the length of edge $(c,g)$ is decreased by $1$, the length of edge $(g,j)$ is increased by $3$, and the weight of edge $(f,i)$ is decreased by 8 \cite{chan2009shortest}}
\label{fluxWithMixture}
\end{figure}

As can be seen in Fig. \ref{fluxWithMixture}, the changing trend of the flux associated with every edge is shown. Several edges such as edge $(h,k)$ disappear from the SPT while edges $(f,i)$ and $(i,n)$ become the elements constructing SPT after the update of edge weight. The other edges such as edges $(k,p)$ and $(b,e)$ are not affected by the update of the edge weight. Fig. \ref{mixture}(b) gives the final SPT after the update of the edge weights. From Fig. \ref{mixture}(b), it can be concluded that the amoeba algorithm can distinguish all the nodes adaptively.


In summary, all the edge updates including the increase, decrease, mixed change of the edge weight can be handled efficiently by the adaptive amoeba algorithm.

\section*{Acknowledgment}
All the authors of the cited papers for providing their network data. The work is partially supported by Chongqing Natural Science Foundation, Grant No. CSCT, 2010BA2003, National Natural Science Foundation of China, Grant No. 61174022, 61364030,
National High Technology Research and Development Program of China (863 Program) (No.2013AA013801), the Fundamental Research Funds for the Central Universities, Grant No. XDJK2012D009.

\begin{table}[htbp]
  \centering
  \caption{The related parameters of the {\em erdos.renyi.game} function. In this function, the parameter $p$ denotes the probability for drawing an edge between two arbitrary vertices.}
    \begin{tabular}{rrrrr}
    \addlinespace
    \toprule
            Dataset &     Number of Nodes   &  $p$ & Number of Edges   \\
    \midrule
    dataset 1 & 500  & 0.0200  & 5000  \\
    dataset 2 & 1000  & 0.0100  & 9921  \\
    dataset 3 & 1500  & 0.0060  & 13475  \\
    dataset 4 & 2000  & 0.0050  & 19903  \\
    \bottomrule
    \end{tabular}%
  \label{dataset}%
\end{table}%

\begin{table}[!ht]\footnotesize
\caption{Adaptive Amoeba Algorithm(L, V, E) for Directed Networks}
\centering
\begin{tabular}{ll}
\toprule
\begin{minipage}{4in}
\begin{algorithmic}
\STATE // $L$ is an $N \times N$ matrix, $L_{ij}$ denotes the length between node $i$ and node $j$
\STATE // $V$ denote the set of nodes, $E$ denotes the set of edges
\STATE // $s$ is the starting node, $t$ is the ending node \\
${D_{ij}} \leftarrow \left( {0,1} \right]\;\left( {\forall i,j = 1,2, \ldots ,N \wedge {L_{ij}} \ne 0} \right)$ \\
${Q_{ij}} \leftarrow 0\;\left( {\forall i,j = 1,2, \ldots ,N} \right)$ \\
${p_i} \leftarrow 0\;\left( {\forall i = 1,2, \ldots ,N} \right)$ \\
$count \leftarrow 1$   \\
\REPEAT
   \STATE ${p_t} \leftarrow 0$  // the pressure at the ending node $t$ \\
   \STATE Calculate the pressure of every node using Eq. (\ref{newFlow})

   \[\sum\limits_{i} {\left( {\frac{{{D_{ij}}}}{{{L_{ij}}}} + \frac{{{D_{ji}}}}{{{L_{ji}}}}} \right)\left( {{p_i} - {p_j}} \right)}  = \left\{ \begin{array}{l}
  + 1\;\;\;\;\;for\;j = s \\
  - 1\;\;\;\;\;for\;j = t \\
 0\;\;\;\;\;\;\;otherwise \\
 \end{array} \right.\]

   \STATE ${Q_{ij}} \leftarrow {{{D_{ij}} \times \left( {{p_i} - {p_j}} \right)} \mathord{\left/
 {\vphantom {{{D_{ij}} \times \left( {{p_i} - {p_j}} \right)} {{L_{ij}}}}} \right.
 \kern-\nulldelimiterspace} {{L_{ij}}}}$ // Using Eq. (\ref{flux})

         \IF {${Q_{ij}} < 0$}
             \STATE ${Q_{ij}} = 0$
         \ENDIF

  \STATE ${D_{ij}} \leftarrow  {Q_{ij}}  + {D_{ij}}$  // Using Eq. (\ref{eql6})
  \STATE $count \leftarrow count + 1$
\UNTIL{a termination criterion is met}
\end{algorithmic}
\end{minipage}\\
\bottomrule
\end{tabular}
\label{algorithm1}
\end{table}

\begin{table}[!ht]\footnotesize
\caption{Adaptive Amoeba Algorithm(L, V, E) for Shortest Path Tree}
\centering
\begin{tabular}{ll}
\toprule
\begin{minipage}{4in}
\begin{algorithmic}
\STATE // $L$ is an $N \times N$ matrix, $L_{ij}$ denotes the length between node $i$ and node $j$
\STATE // $V$ denote the set of nodes, $E$ denotes the set of edges
\STATE // $s$ is the root node \\
${D_{ij}} \leftarrow \left( {0,1} \right]\;\left( {\forall i,j = 1,2, \ldots ,N \wedge {L_{ij}} \ne 0} \right)$ \\
${Q_{ij}} \leftarrow 0\;\left( {\forall i,j = 1,2, \ldots ,N} \right)$ \\
${p_i} \leftarrow 0\;\left( {\forall i = 1,2, \ldots ,N} \right)$ \\
$count \leftarrow 1$   \\
\REPEAT
   \STATE Calculate the pressure of every node using Eq. (\ref{SPTFlow})

   \[\sum\limits_{i} {\left( {\frac{{{D_{ij}}}}{{{L_{ij}}}} + \frac{{{D_{ji}}}}{{{L_{ji}}}}} \right)} \left( {{p_i} - {p_j}} \right) = \left\{ \begin{array}{l}
  + 1\quad \;for\;j = s \\
 \frac{-1}{{N - 1}}\;\;\;for\;j \ne s\quad  \\
 \end{array} \right.\]

   \STATE ${Q_{ij}} \leftarrow {{{D_{ij}} \times \left( {{p_i} - {p_j}} \right)} \mathord{\left/
 {\vphantom {{{D_{ij}} \times \left( {{p_i} - {p_j}} \right)} {{L_{ij}}}}} \right.
 \kern-\nulldelimiterspace} {{L_{ij}}}}$ // Using Eq. (\ref{flux})

         \IF {${Q_{ij}} < 0$}
             \STATE ${Q_{ij}} = 0$
         \ENDIF

  \STATE ${D_{ij}} \leftarrow  {Q_{ij}}  + {D_{ij}}$  // Using Eq. (\ref{eql6})
  \STATE $count \leftarrow count + 1$
\UNTIL{a termination criterion is met}
\end{algorithmic}
\end{minipage}\\
\bottomrule
\end{tabular}
\label{algorithm2}
\end{table}

\section*{Author contributions}
 X. Z., Q. L., Z. Z. designed and performed research. X. Z. wrote the paper. Y. H. performed the computation. S. M., F. T.S. C. and Y. D. analyzed the data. All authors discussed the results and commented on the manuscript.

\section*{Additional information}
\textbf{Competing financial interests:} The authors declare no competing financial interests.

\end{document}